\documentclass[sigconf]{acmart}

\AtBeginDocument{%
  }

\copyrightyear{2024} 
\acmYear{2024} 
\setcopyright{acmlicensed}\acmConference[ISWC '24]{Proceedings of the 2024 ACM International Symposium on Wearable Computers}{October 5--9, 2024}{Melbourne, VIC, Australia}
\acmBooktitle{Proceedings of the 2024 ACM International Symposium on Wearable Computers (ISWC '24), October 5--9, 2024, Melbourne, VIC, Australia}
\acmDOI{10.1145/3675095.3676622}
\acmISBN{979-8-4007-1059-9/24/10}

\usepackage{algorithmic}
\usepackage{graphicx}
\usepackage{multirow}
\usepackage{booktabs}
\usepackage{siunitx}
\usepackage[justification=centering]{caption}
\usepackage{url}
\usepackage{colortbl}
\usepackage{tabularx}
\usepackage{textcomp}
\usepackage{xcolor}
\usepackage{xspace}

\begin{document}

\title{WorkR: Occupation Inference for Intelligent Task Assistance}


\author{Yonchanok Khaokaew}
\orcid{0000-0003-4297-6274}
\email{y.khaokaew@unsw.edu.au}
\author{Hao Xue}
\orcid{0000-0003-1700-9215}
\email{hao.xue1@unsw.edu.au}
\affiliation{%
  \institution{School of Computer Science and Engineering,\\ University of New South Wales }
    \city{Sydney}
  \state{NSW}
  \country{Australia}
}

\author{Mohammad Saiedur Rahaman}
\orcid{0000-0003-2320-0112}
\email{m.saiedurrahaman@cqu.edu.au}
\affiliation{%
  \institution{School of Engineering and Technology,\\
  CQUniversity}
    \city{Melbourne}
  \state{VIC}
  \country{Australia}
}

\author{Flora D. Salim}
\orcid{0000-0002-1237-1664}
\email{flora.salim@unsw.edu.au}
\affiliation{%
  \institution{School of Computer Science and Engineering, \\University of New South Wales }
    \city{Sydney}
  \state{NSW}
  \country{Australia}
}

\renewcommand{\shortauthors}{Khaokaew et al.}


\begin{abstract}

Occupation information can be utilized by digital assistants to provide occupation-specific personalized task support, including interruption management, task planning, and recommendations. Prior research in the digital workplace assistant domain requires users to input their occupation information for effective support. However, as many individuals switch between multiple occupations daily, current solutions falter without continuous user input. To address this, this study introduces \textit{WorkR}, a framework that leverages passive sensing to capture pervasive signals from various task activities, addressing three challenges: the lack of a passive sensing architecture, personalization of occupation characteristics, and discovering latent relationships among occupation variables. We argue that signals from application usage, movements, social interactions, and the environment can inform a user's occupation. \textit{WorkR} uses a Variational Autoencoder (VAE) to derive latent features for training models to infer occupations. Our experiments with an anonymized, context-rich activity and task log dataset demonstrate that our models can accurately infer occupations with more than 91\% accuracy across six ISO occupation categories.

\end{abstract}

\ccsdesc[500]{Computing methodologies~Classification and regression trees}
\ccsdesc[500]{Human-centered computing~Empirical studies in ubiquitous and mobile computing}
\ccsdesc[300]{Human-centered computing~Ubiquitous computing}
\ccsdesc[300]{Computing methodologies~Modeling methodologies}

\keywords{Behaviour Modelling, Occupation Inference, Digital Assistant, Ubiquitous and Mobile Computing}

\received{15 May 2024}
\received[revised]{23 July 2024}
\received[accepted]{26 July 2024}


\maketitle

\section{Introduction}\label{section:intro}
\begin{figure}[ht!]
\vspace{-10pt}
  \centering
  \includegraphics[width=0.85\linewidth]{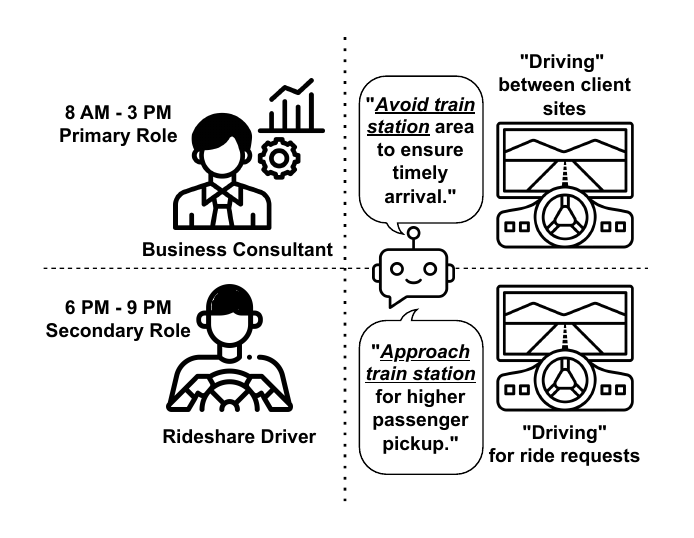}
  \vspace{-20pt}
  \caption{Motivation for Adaptive Digital Assistance in Occupational Contexts}
  \Description{Motivation for Adaptive Digital Assistance in Occupational Contexts}
  \label{fig:job_challenge}
  \vspace{-5pt}
\end{figure}

Digital assistants (DAs) have become essential tools for enhancing productivity among knowledge workers by managing tasks and schedules with precision. Inspired by prior work that highlights the varying support needs across different professional roles \cite{myers2007intelligent, sun2016intelligent, khaokaew2022imagining}, we observe that even when performing identical tasks such as attending a meeting, individuals may require distinct support from DAs. For example, in a meeting scenario involving a salesperson presenting, a manager deciding on a purchase, a secretary tasked with record-keeping, and an ICT specialist providing technical support, each participant benefits differently from DA functionalities. The salesperson might need all personal notifications silenced, the manager could benefit from real-time data verification of the presentation content, the secretary may want automated key point logging, and the ICT specialist might prefer instant access to relevant technical details.

\begin{table*}[ht!]
\footnotesize
\centering

\caption{Participants and Their Occupation Groups Based on the International Standard Classification of Occupations (ISCO) }
\label{tab:professionals}
\resizebox{0.98\textwidth}{!}{\begin{tabular}{ p{12cm} c }
\toprule
\textbf {Occupation | Occupation Details} & \textbf{Number of annotations} \\
\toprule
\textbf{Professionals }--Library officer, Structural engineer, Business analyst, Marketing consultant, Technical writer and editor, Principal advisor, Financial Specialist, Book editor, Artist, Job recruiter, Graphic designer, Tutor &  944 \\ \midrule
\textbf{Managers }--Business development manager, Strategy manager, Start-up founder, Student service, Business owner, Event manager, Account manager, Product owner, Sales manager, Project manager, Career counsellor &  849 \\ \midrule
\textbf{ICT professional }--Development relation specialist, Software delivery lead, Software developer, It security analyst, Digital service analyst, Data specialist, Ux designer, Test analyst &  554 \\ \midrule
\textbf{Student }--University student &  358 \\ \midrule
\textbf{Technicians and associate professionals }--Project coordinator, Lab technician, Administrator/program assistant, Research assistant, Public health internship &  197 \\ \midrule
\textbf{Service and sales workers }--Sales and marketing, Chef, University Bartender & 148 \\ 
\bottomrule

\end{tabular}}

\end{table*}

Furthermore, the rise in multiple job holders—from 5-6\% to 6-7\% post-COVID \cite{MJHABS2023}—illustrates a dynamic workforce where individuals frequently switch roles throughout the day, further complicating the traditional operation of DAs. It becomes impractical and intrusive for DAs to constantly request users to update their occupation or context.
Given these challenges, this study poses a critical question: \textit{\textbf{"Can we infer individuals’ occupations using passive sensor logs associated with their tasks?"}} Addressing this question could transform how DAs understand and interact with users, by aligning support mechanisms more closely with their immediate professional demands without intrusive queries. Advances in sensory and behavioural data mining have introduced promising avenues for passively collecting such data, facilitating efficient task management and enabling systems to predict and adapt to user autonomously needs \cite{graus2016analyzing, actrecpercom}.

To tackle these complexities, our study introduces \textit{WorkR}, a novel framework that leverages multi-source sensor fusion and machine learning to infer occupations from passive sensor logs. This framework aims to enhance the functionality of DAs by enabling more context-aware interactions that could potentially operate within the users' devices, minimizing the need for direct data querying. While not directly addressing user privacy at this stage, the localized processing of data suggests a path towards reducing privacy risks in future implementations. 

The \textit{WorkR} framework utilizes a variational autoencoder to derive latent space representations of sensor data, which are then processed through an XGBoost model to predict occupations accurately \cite{belkadi2020intelligent, Sarker2018}. This method aims to provide DAs with deeper insights into the user’s occupational context, potentially improving the assistance they provide. Our key contributions are as follows: 

\begin{enumerate}
    \item Introduction of the occupation inference problem as a new challenge for passive sensing and machine learning, with potential applications in enhancing digital assistant functionalities.
    \item Development of \textit{WorkR}, a framework that utilizes multi-sensor logs and machine learning to infer occupations without active user input, focuses on preserving user privacy through anonymized data.
    \item Utilization of a variational autoencoder to learn latent factors from multi-source sensor data associated with user tasks, enhancing the predictive accuracy of the occupation inference.
\end{enumerate}
    
\textit{WorkR} differentiates itself by eliminating the need for active user participation and using anonymized datasets to ensure privacy. This approach aims to pave the way for future digital assistants that can offer highly personalized support based on dynamically inferred occupational data, significantly enhancing both functionality and user engagement. The paper will further discuss the dataset, the implementation of the \textit{WorkR} model, experimental evaluations, and the implications of our findings.

\section{Activity and Task Log Dataset and Analysis }\label{section:dataset}

This study utilizes an anonymized task and activity log dataset collected from 53 participants over one year between May 2018 and June 2019 \cite{liono2019building, liono2020intelligent, khaokaew2022imagining}. All participants, fluent in English and residing in Australia, are diverse in terms of occupations. We categorize the occupation group of these participants by using the major groups from the International Standard Classification of Occupations (ISCO) \cite{hoffmann2003international}. These major groups found in our dataset consist of \textit{Managers},	
\textit{Professionals}, \textit{Service and sales workers}, \textit{Technicians and associate professionals}. Then, we further separate users in the Information and Communications Technology Professional (ICT Professional) group, which is the minor group in the Professional groups since almost half of the users from the Professional group work in this field, and the characteristics and tasks performed by \textit{ICT professional} participants are different from other Professionals. In addition, we also consider the \textit{Student} as a part of the participant's occupation in this study. Note that although some participants have multiple jobs due to the limited number, we will only consider the primary job in this work.

The dataset comprises \textbf{sensor signals} and manually logged \textbf{task annotations} from users’ mobile and desktop devices. Sensor signals were passively logged through three applications: an in-house developed app, RescueTime, and Journeys, capturing data in a timestamped manner to ensure precision and relevance to the work-related tasks recorded. Task annotations were manually collected using the Experience Sampling Method (ESM) via the in-house developed app. ESM provided in-situ annotations through brief, non-intrusive surveys focused on tasks performed within the last hour, thereby reducing disruption and cognitive biases \cite{hektner2007experience}. These surveys captured details about the task description, task category, and whether the task was work-related. Weekly interviews with participants were conducted to confirm the accuracy of these task annotations, ensuring robust data collection. Task annotations were paired with sensor data based on timestamps, allowing for detailed analysis of the contextual and behavioral patterns associated with specific occupational activities. In detailing the types of sensor signals utilized, we categorized them into four groups, each reflecting a different dimension of user activity and environment:

\textbf{Physical features ($P$)}: Captured through smartphone sensors, these include mean accelerometer readings across three axes, averages from gyroscope and magnetometer sensors to track motion and orientation, and the average number of steps per time slot, along with data on visited locations.

\textbf{Application Usage Features ($A$)}: Derived from users' smartphone app usage logs, detailing categorized app usage patterns and screen interaction dynamics. This includes an analysis of app usage time distribution across categories and the proportion of screen-on time within each period.

\textbf{Social and Environmental Features ($S$)}: Insights from smartphone sensors about users' surroundings and social interactions. This includes noise level statistics and proximity to others through counts of nearby Bluetooth devices and WiFi access points and detailed environmental data such as ambient air pressure from the barometer sensor.

\textbf{Temporal Features ($T$)}: Focusing on the timing aspects of users' activities, utilizing timestamps to determine the specific hours of work-related tasks.

These sensor data groups collectively enable a comprehensive analysis of occupational behaviours and environmental contexts. Building on this foundation, we established ground truth for each participant's job on an hourly basis. We selected periods mentioned by users as work-related and where the task description matched their primary job. Out of the total, 46 participants provided sufficient work-related data (with more than two weeks of data and all types of sensors) that were used for analysis in this article. The occupation distribution in our dataset is detailed in Table \ref{tab:professionals}. Building on this data foundation, the next subsection conducts an exploratory analysis, examining task similarities and sensor signals across different occupations to identify distinctive and overlapping occupational activities.

\begin{table*}[h!]
\footnotesize
  \centering
  \caption{Top 10 Application Usage by Occupation: Proportional Distribution Across Categories. This table shows the top 10 application categories used, segmented by occupation. ( Comm. = Communication, Ref. = Reference, Edu. = Education, Mang. = Management, P\&V = Photo \& Video)}  \vspace{-8pt}\resizebox{0.89\textwidth}{!}{\begin{tabular}{ccccccccccc}

    \toprule

\textbf{ Occupation } &\textbf{ Comm. } &\textbf{ Social } &\textbf{ Ref. } &\textbf{ P\&V } &\textbf{ Shopping } &\textbf{ Edu. } &\textbf{ Finance } &\textbf{ Manag } &\textbf{ Music } &\textbf{ Games } \\ 
\toprule
PROFESSIONALS &\cellcolor[rgb]{   0.522 , 0.784 , 0.49 } 0.19 &\cellcolor[rgb]{   0.984 , 0.918 , 0.518 } 0.13 &\cellcolor[rgb]{   0.722 , 0.843 , 0.502 } 0.15 &\cellcolor[rgb]{   0.722 , 0.843 , 0.502 } 0.15 &\cellcolor[rgb]{   0.722 , 0.843 , 0.502 } 0.16 &\cellcolor[rgb]{   0.976 , 0.553 , 0.447 } 0.08 &\cellcolor[rgb]{   0.973 , 0.412 , 0.42 } 0.05 &\cellcolor[rgb]{   0.973 , 0.412 , 0.42 } 0.05 &\cellcolor[rgb]{   0.973 , 0.412 , 0.42 } 0.02 &\cellcolor[rgb]{   0.973 , 0.412 , 0.42 } 0.02 \\  \hline
MANAGERS &\cellcolor[rgb]{   0.522 , 0.784 , 0.49 } 0.19 &\cellcolor[rgb]{   0.522 , 0.784 , 0.49 } 0.20 &\cellcolor[rgb]{   0.984 , 0.918 , 0.518 } 0.14 &\cellcolor[rgb]{   0.984 , 0.918 , 0.518 } 0.13 &\cellcolor[rgb]{   0.984 , 0.918 , 0.518 } 0.08 &\cellcolor[rgb]{   0.976 , 0.553 , 0.447 } 0.08 &\cellcolor[rgb]{   0.976 , 0.553 , 0.447 } 0.06 &\cellcolor[rgb]{   0.976 , 0.553 , 0.447 } 0.06 &\cellcolor[rgb]{   0.973 , 0.412 , 0.42 } 0.04 &\cellcolor[rgb]{   0.973 , 0.412 , 0.42 } 0.04 \\  \hline
ICT &\cellcolor[rgb]{   0.388 , 0.745 , 0.482 } 0.23 &\cellcolor[rgb]{   0.984 , 0.918 , 0.518 } 0.13 &\cellcolor[rgb]{   0.522 , 0.784 , 0.49 } 0.17 &\cellcolor[rgb]{   0.984 , 0.918 , 0.518 } 0.13 &\cellcolor[rgb]{   0.976 , 0.553 , 0.447 } 0.06 &\cellcolor[rgb]{   0.984 , 0.918 , 0.518 } 0.08 &\cellcolor[rgb]{   0.973 , 0.412 , 0.42 } 0.05 &\cellcolor[rgb]{   0.973 , 0.412 , 0.42 } 0.02 &\cellcolor[rgb]{   0.973 , 0.412 , 0.42 } 0.05 &\cellcolor[rgb]{   0.976 , 0.553 , 0.447 } 0.07 \\  \hline
STUDENT &\cellcolor[rgb]{   0.722 , 0.843 , 0.502 } 0.15 &\cellcolor[rgb]{   0.722 , 0.843 , 0.502 } 0.14 &\cellcolor[rgb]{   0.522 , 0.784 , 0.49 } 0.17 &\cellcolor[rgb]{   0.722 , 0.843 , 0.502 } 0.16 &\cellcolor[rgb]{   0.984 , 0.918 , 0.518 } 0.10 &\cellcolor[rgb]{   0.984 , 0.918 , 0.518 } 0.08 &\cellcolor[rgb]{   0.973 , 0.412 , 0.42 } 0.03 &\cellcolor[rgb]{   0.973 , 0.412 , 0.42 } 0.03 &\cellcolor[rgb]{   0.976 , 0.553 , 0.447 } 0.08 &\cellcolor[rgb]{   0.976 , 0.553 , 0.447 } 0.06 \\  \hline
SERVICE &\cellcolor[rgb]{   0.388 , 0.745 , 0.482 } 0.23 &\cellcolor[rgb]{   0.522 , 0.784 , 0.49 } 0.17 &\cellcolor[rgb]{   0.984 , 0.918 , 0.518 } 0.11 &\cellcolor[rgb]{   0.522 , 0.784 , 0.49 } 0.17 &\cellcolor[rgb]{   0.976 , 0.553 , 0.447 } 0.07 &\cellcolor[rgb]{   0.976 , 0.553 , 0.447 } 0.06 &\cellcolor[rgb]{   0.973 , 0.412 , 0.42 } 0.03 &\cellcolor[rgb]{   0.976 , 0.553 , 0.447 } 0.06 &\cellcolor[rgb]{   0.973 , 0.412 , 0.42 } 0.03 &\cellcolor[rgb]{   0.976 , 0.553 , 0.447 } 0.06 \\  \hline
TECHNICIANS &\cellcolor[rgb]{   0.984 , 0.918 , 0.518 } 0.13 &\cellcolor[rgb]{   0.522 , 0.784 , 0.49 } 0.17 &\cellcolor[rgb]{   0.722 , 0.843 , 0.502 } 0.15 &\cellcolor[rgb]{   0.722 , 0.843 , 0.502 } 0.16 &\cellcolor[rgb]{   0.984 , 0.918 , 0.518 } 0.10 &\cellcolor[rgb]{   0.976 , 0.553 , 0.447 } 0.07 &\cellcolor[rgb]{   0.976 , 0.553 , 0.447 } 0.08 &\cellcolor[rgb]{   0.976 , 0.553 , 0.447 } 0.06 &\cellcolor[rgb]{   0.976 , 0.553 , 0.447 } 0.06 &\cellcolor[rgb]{   0.973 , 0.412 , 0.42 } 0.03 \\  \hline

    \end{tabular}}
  \label{tab:Application freq diff}%
\end{table*}%

\subsection{Task similarity}

\begin{figure}[h!]

  \centering
  \includegraphics[scale=0.45]{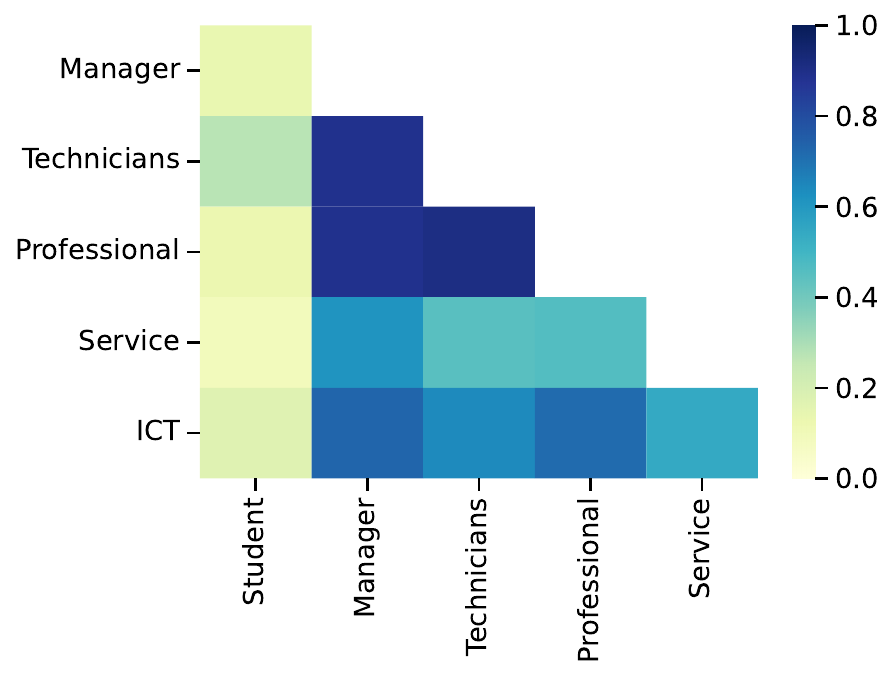}
  \vspace{-5pt}
  \caption{The figure shows the cosine similarity across different occupations based on their tasks}
    \Description{The figure shows the cosine similarity across different occupations based on their tasks }
  \label{fig:job_similarity}
    \vspace{-5pt}
\end{figure}

We began by examining task category proportions reported by participants, totalling 6,321 tasks, from which we selected 3,050 work-related instances for further analysis. To explore how task distributions differ among various occupations, we conducted a task similarity analysis across six occupational groups: Student, Manager, Technician, Professional, Service, and ICT. For each occupation, we constructed a task frequency vector and computed cosine similarities between the vectors of all occupations. Figure \ref{fig:job_similarity} displays a task similarity matrix, illustrating the relationships across these groups. The heatmap indicates significant variations in task profiles among different occupations, reflecting the unique nature of tasks performed by each occupational group. While some occupations show moderate similarities in task distributions, suggesting overlapping responsibilities, others differ markedly, highlighting the diversity of work-related activities among different roles. This variation emphasizes the challenge of inferring occupations based solely on task features, as overlapping tasks can lead to ambiguity in occupation identification. displays a task similarity matrix, illustrating the relationships across these groups.

\subsection{Application usage information}

Application usage patterns, while reflective of occupational activities, can also reveal personal or secondary activities that are not directly job-related but are still prevalent among certain professional groups. This observation underscores the complexity of using app usage data for occupation inference and highlights the importance of contextual understanding.

For instance, while managers commonly use messaging and social networking apps like Slack, Facebook, and LinkedIn, which directly support their roles in communication and networking, service workers frequently use communication apps (e.g., WhatsApp, dialer) alongside photo\& video apps (e.g., Camera, YouTube). However, these apps could be used for purposes like documenting work-related activities or sharing visual content pertinent to customer service tasks, which are essential but less obviously tied to their primary job functions. Our analysis revealed that app usage varies subtly across different occupations (Table. \ref{tab:Application freq diff}). Managers predominantly utilize apps that facilitate professional networking and communication, whereas service workers employ a mix of tools that blend professional utility with personal usage. This mixed usage pattern suggests that some apps, though not exclusively work-related, are integral to the daily routines of specific occupational groups.

\subsection{Physical activities information}
\vspace{-4pt}
\begin{figure}[h!]
  \centering
  \includegraphics[width=\linewidth]{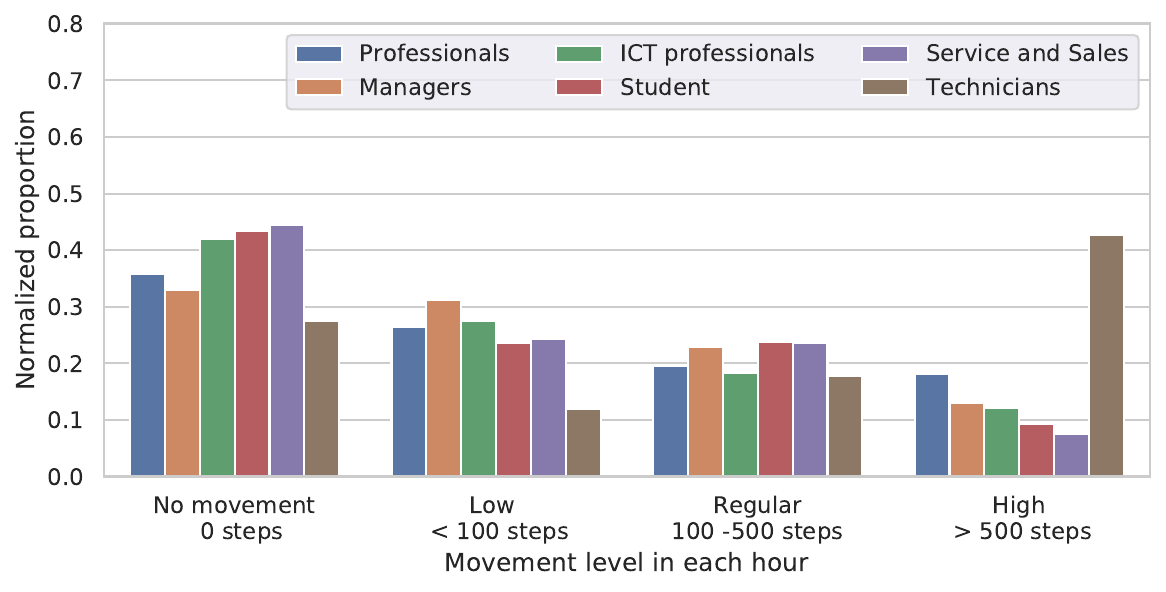}
 \vspace{-15pt}
  \caption{The number of steps across different occupations }
  \Description{The number of steps across different occupations }
  \label{fig:movements}
  
\end{figure}

To analyze physical activity similarities, we compared hourly steps recorded by the Sensing app across various occupations, as detailed in Section \ref{section:dataset}. Figure \ref{fig:movements}  shows that Students, ICT professionals, and Service \& Sales workers generally exhibit low movement levels per hour. Students and ICT professionals often stay in one place while working, whereas service and sales workers, such as chefs and bartenders, rarely actively use their devices during work hours. This detail was confirmed through weekly interviews with participants. The Technician group stands out, with over 40 percent exhibiting high movement levels (more than 500 steps per hour), compared to just 10-20 percent in other groups. While movement data alone may not suffice to determine occupation, it is a valuable feature that can improve the model's accuracy.

\subsection{Social and Environmental information}
Environmental factors significantly aid in identifying the working conditions of different occupational groups. For instance, Managers or Project Coordinators typically work in socially dense environments, while Factory Technicians might work in areas with high noise levels.

\begin{figure}[h!]
  \centering
  \includegraphics[width=\linewidth]{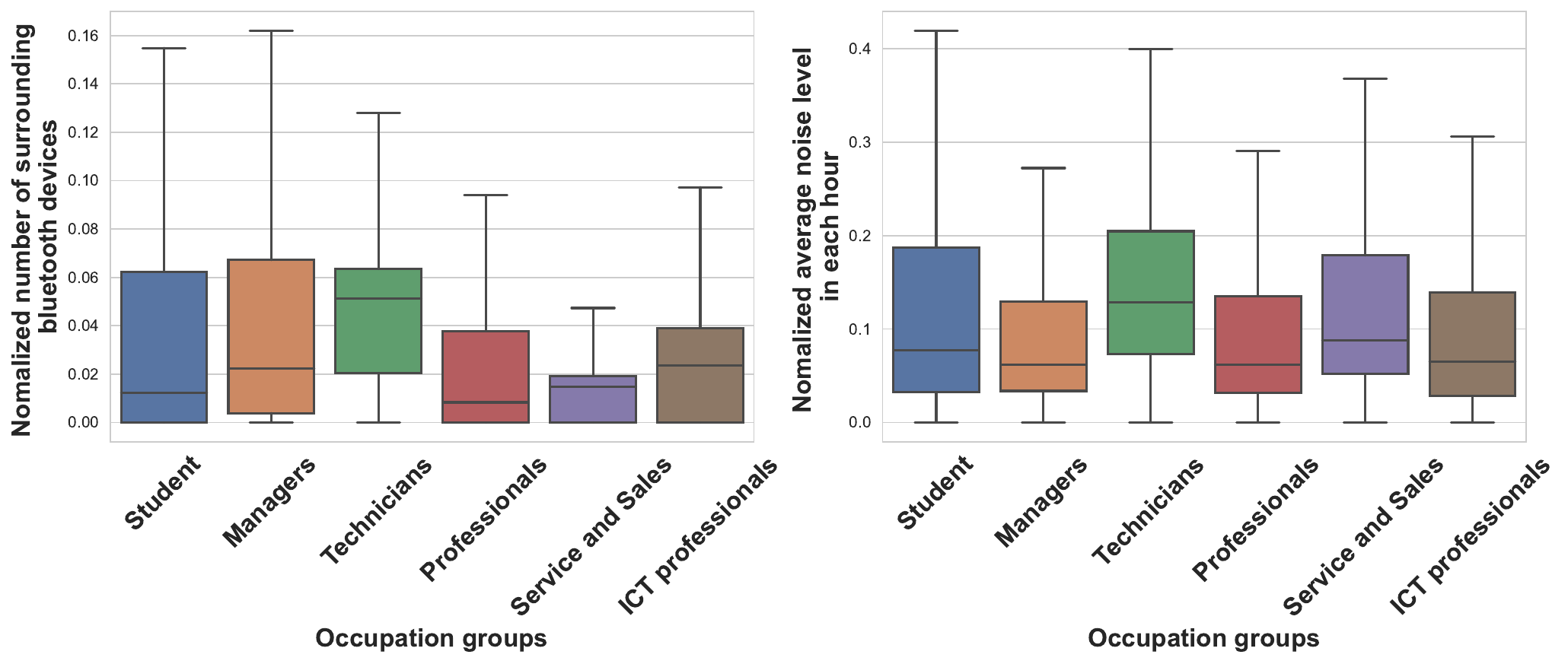}
  \caption{Distribution of surrounding Bluetooth devices (left), and  the surrounding noise level (right) }
  \Description{Distribution of surrounding Bluetooth devices (left), and  the surrounding noise level (right)}
  \label{fig:EVC}
\vspace{-5px}
\end{figure}

Our dataset includes social and environmental data captured via noise and Bluetooth sensors on participants' smartphones. Bluetooth scans measured the presence of nearby devices, indicating social density, while noise sensors recorded ambient sound levels during tasks. As shown in Figure \ref{fig:EVC}, Technicians often work in noisier settings, likely due to increased human activity. In contrast, roles in Service and Sales experience high noise levels despite fewer social interactions, underscoring the need to consider both social and environmental factors for accurate occupational inference.

\subsection{Temporal information}
Temporal information is crucial in various fields and has been shown to be effective in predicting different patterns of activity \cite{krismayer2019predicting, tahmasbi2018modeling}. In the context of occupational tasks, time can be a defining factor as different jobs often perform similar tasks at different times of the day. For instance, Bartenders typically begin their shifts in the evening, unlike most other professions, which start in the morning.

\begin{figure}[h!]
  \centering
  \includegraphics[width=\linewidth]{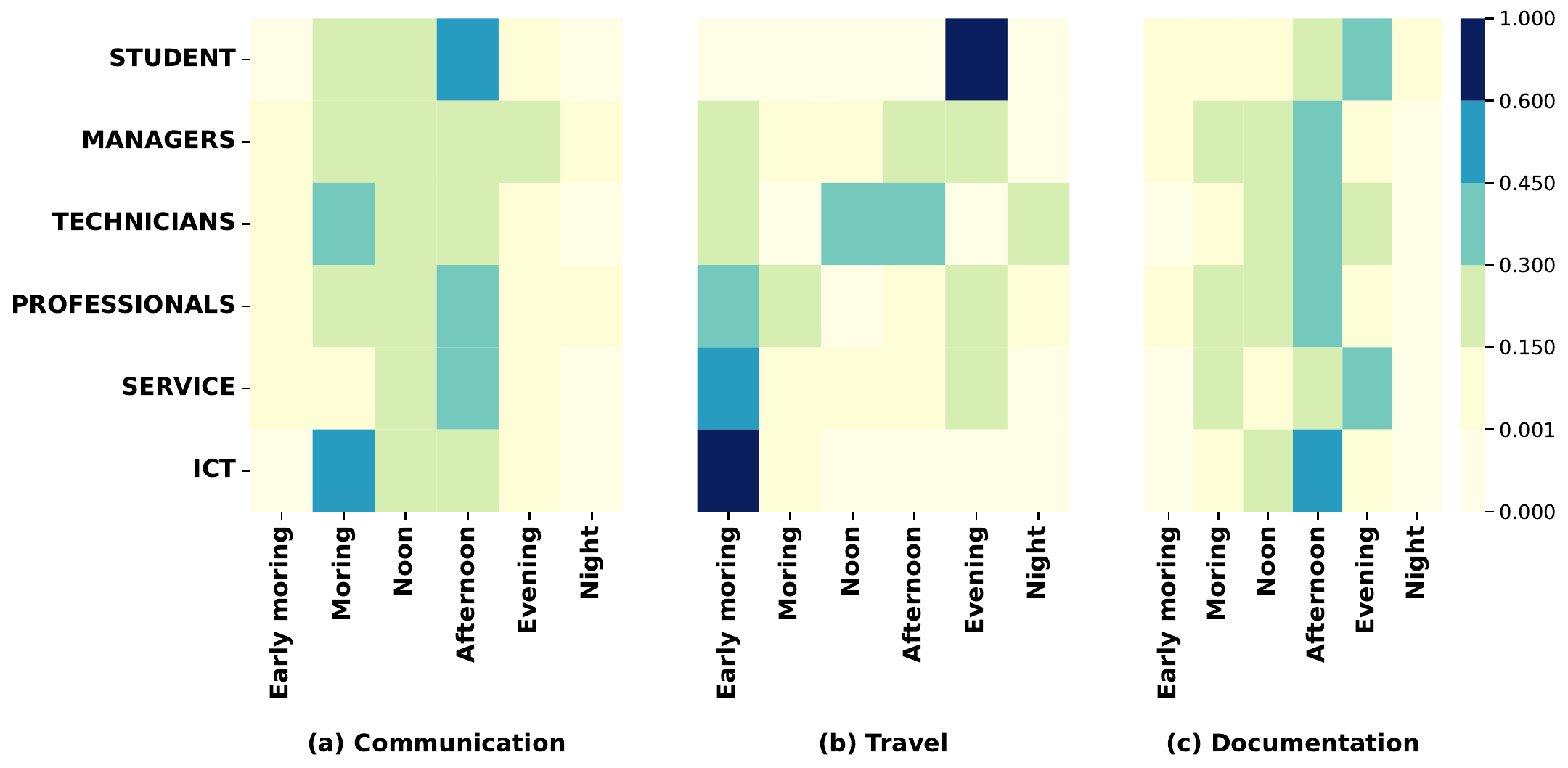}
  \caption{The heatmap of 3 general tasks performed across different occupations on each temporal segment}
  \Description{The heatmap of 3 general tasks performed across different occupations on each temporal segment}
  \label{fig:heatmap_task_frequiencies}
  \vspace{-8pt}
\end{figure}

To validate the importance of temporal patterns, we analyzed the distribution of general tasks like communication, travel, and documentation across occupations, segmenting task occurrences into six parts of the day: early morning, morning, noon, afternoon, evening, and night. Figure \ref{fig:heatmap_task_frequiencies} illustrates distinct temporal task patterns among occupations. Notably, Service \& Sales workers and Students often complete documentation tasks in the evening, contrasting with other occupations, which tend to do so in the afternoon. These findings highlight the potential of temporal information as a valuable feature for inferring occupations based on task timing.

\section{Occupation Inference}

This section defines the occupation inference problem, discusses the features used, and presents our occupation inference model.

\subsection{Problem Definition}

Assume we have an occupation of interest denoted as \( J \), which we aim to infer for a collection of users. For this purpose, we train a multi-class classifier. Given a set of signals associated with a test instance \( x \) in time slot \( t \), the trained classifier infers the corresponding occupation of \( x \) in this time slot, denoted as \( J_x \), as shown below:
\begin{equation}
    g(F^t_{p}, F^t_{l}) \rightarrow J_x
\end{equation}
where \( g(\cdot) \) is a function that maps the combination of processed features \( F^t_p \) and latent features \( F^t_l \) collected at time slot $t$ to the occupations listed in Table \ref{tab:professionals}.

\subsection{Feature extraction } \label{subsection:features}

The features for our analysis were constructed from raw logs collected from users' smartphones and desktops, using data recorded during tasks and from sensor apps. Each feature was computed and generated within time slots of 900 seconds (15 minutes), applying a sliding window approach. Categorical data, such as the weekday and hour of the day, were transformed using one-hot encoding to capture temporal patterns effectively. Application usage was quantified by calculating the ratio of each app category used and the screen's duration within each time slot. Additionally, we normalized various data types to a range of [0,1] using the min-max normalization technique to ensure uniformity and comparability. This included statistical measurements from IMU sensors and barometers (mean, median, standard deviation, maximum, minimum, interquartile range, and root mean square), noise level measurements (mean, maximum, and minimum values), and count data such as the number of visited locations, Bluetooth and WiFi access points, and steps taken. This feature extraction approach ensures that our models receive well-prepared and standardized input, enhancing the reliability of our findings.

\subsection{WorkR: occupation Inference model}
\vspace{-10pt}
\begin{figure}[h!]
  \centering
  \includegraphics[width=0.90\linewidth]{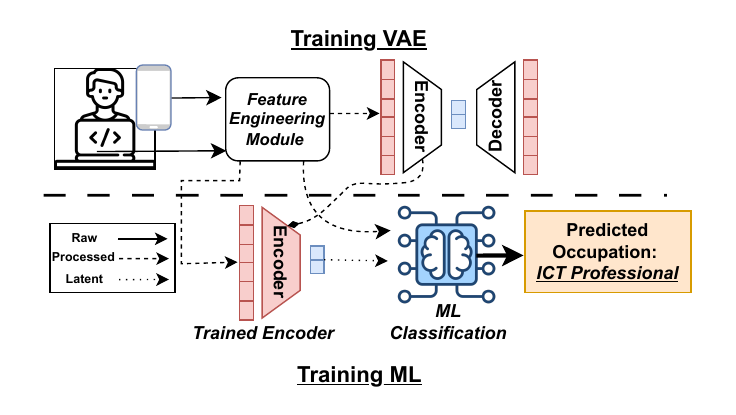}
  \vspace{-15pt}
  \caption{An overview of our proposed model}
  \Description{An overview of our proposed model}
   \label{fig:systemOverview}
   \vspace{-5pt}
\end{figure}

We develop a model called \textit{WorkR} to infer occupations. The overview of our model is shown in Figure. \ref{fig:systemOverview}. Firstly, the raw features associated with a particular occupation's tasks are collected and utilised as input to the \textit{WorkR} model. In the next step, raw signals are aggregated and undergo temporal segmentation for feature engineering. This process involves constructing a set of hand-made features, as discussed in the prior section. Like any pervasive sensing application with numerous features, identifying latent relationships among variables is challenging. Hence, we employ a latent feature construction engine, the core part of \textit{WorkR}. 

We apply Variational Autoencoders (VAEs) \cite{kingma2013auto}, a technique widely used in various studies \cite{wang2019extracting,latif2017variational}, to extract latent information from raw features. We chose VAEs because they regularize the distribution of encodings during training, preventing overfitting and ensuring that the latent space possesses desirable properties. Our VAE implementation is adapted from a Keras tutorial \footnote{https://blog.keras.io/building-autoencoders-in-keras.html}. We tested latent dimensions ranging from 2 to 32, finding that a dimension of 20 provided optimal performance. After training the VAE model with the preprocessed features, we use its encoder to generate latent features. Combined with the preprocessed features, these latent features are then used as input to train machine learning models for inferring users' occupations in the current time slot. We will discuss the results in subsequent sections.

\section{Experiments and Results}

\subsection{Experiment setting}
One of our main endeavours is to classify occupations using the mobile sensing features, as discussed in Section \ref{subsection:features}. We aim to classify the user's occupation by combining the observed sensing features and their latent features within a time slot.  To this end, we implement several different classification models including Support Vector Machine, Naive Bayes~\cite{murphy2006naive}, Multi-layer Perceptron \cite{glorot2010understandingMLP}, random forest \cite{biau2012randomforest}, and Gradient Boosting. To apply our gradient boosting technique, we rely on XGBoost classifiers \cite{chen2016xgboost} since XGBoost is a highly optimised distributed gradient boosting library that is extremely flexible, scalable, and portable. We set the maximum depth parameter to six and the minimum child weight to one for our gradient-boosting model, which achieved optimal performance with these settings. We evaluate the performance of various classification models using combinations of latent and pre-processed features. For the RF model, we configured 100 trees with `balanced' class weight to manage imbalanced data. The MLP was set with 10 hidden layers and limited to 300 iterations for convergence. The SVM used an RBF kernel with both gamma and C set to 1, and a smoothing parameter of 1e-09 was applied for the NB model. We focus on results from latent features derived from physical activity, which showed the most promising outcomes. Details on the ablation study of features will be discussed subsequently.

In order to evaluate our proposed model, we divide the data for each user into three parts. The first 70\% is used for training, the next 10\% for validation, and the final 20\% for testing, all in chronological order. This method ensures no information overlap between the training, validation, and test sets. During the tuning phase, we rely on the accuracy metric computed on the validation set to select the optimal hyperparameters for our model.

\subsection{Experiment Results}

\begin{table}[h!]
\small
\vspace{-6px}
\caption{ WorkR Performance Metrics Across Different ML Models Using Combined Latent and Preprocessed Features}
\vspace{-2px}
\centering
\label{tab:result}
\resizebox{\linewidth}{!}{\begin{tabular}{c|c|c|c|c}
\toprule

\textbf{ML models}&\textbf{F1 score} &\textbf{Precision}&\textbf{Recall} &\textbf{Accuracy} \\


\midrule

NB& 0.2345 $\pm$ 0.008&0.4717 $\pm$ 0.019&0.3900 $\pm$ 0.013&0.2534 $\pm$ 0.009 \\
SVM& 0.3328 $\pm$ 0.012&0.3428 $\pm$ 0.024&0.2168 $\pm$ 0.016&0.2105 $\pm$ 0.021 \\
MLP& 0.6087 $\pm$ 0.034&0.6195 $\pm$ 0.051&0.5576 $\pm$ 0.041&0.5751 $\pm$ 0.042 \\
RF& 0.8838 $\pm$ 0.020&0.8964 $\pm$ 0.004&0.8389 $\pm$ 0.033&0.8623 $\pm$ 0.023 \\
XGBoost& \textbf{0.9193 $\pm$ 0.003}&\textbf{0.9301 $\pm$ 0.003}&\textbf{0.8990 $\pm$ 0.004}&\textbf{0.9118 $\pm$ 0.003} \\

\bottomrule
\end{tabular}}
\vspace{-4px}
\end{table}

We evaluated the performance of various classification techniques and their ability to utilise latent features derived from VAEs. These models' comparative performance has been summarised in Table \ref{tab:result}. XGBoost proved to be the most effective, achieving an accuracy and F1 score of 0.91 each when utilizing a combination of latent and preprocessed features. Random Forest, another tree-based technique, followed with an F1 score of 0.88. In contrast, the MLP scored only 0.60, owing to the limited size of our dataset, which curtails the effectiveness of deep learning models that typically necessitate large volumes of data to perform well.

To further validate our model's robustness, we conducted an ablation study using different combinations of feature sets, the results of which are detailed in Table \ref{tab:abalation_feature}. Utilising the XGBoost model for this analysis, we noted that the combination of physical features, application usage, and social \& environmental features yielded the best performance. This feature set is congruent with the ones used in related work to recognise tasks performed by users \cite{liono2020intelligent, 10.1145/3397271.3401441}. It is interesting to note that as per our findings, temporal information does not significantly contribute to our occupation inference model's accuracy, suggesting that the predictive power of temporal features may be limited in this context.

\begin{table}[ht!]
\small
  \vspace{-3.5pt}
\caption{Ablation Study on the Impact of Different Sets of Preprocessed Features on Model Performance}
  \vspace{-2pt}
\centering
\label{tab:abalation_feature}
\resizebox{0.95\linewidth}{!}{\begin{tabular}{c|c|c|c|c}
\toprule

\textbf{Features}&\textbf{F1 score} &\textbf{Precision}&\textbf{Recall} &\textbf{Accuracy} \\

\midrule

P& 0.8677 $\pm$ 0.003&0.8742 $\pm$ 0.004&0.8628 $\pm$ 0.003&0.8878 $\pm$ 0.003 \\
A& 0.3326 $\pm$ 0.006&0.3742 $\pm$ 0.009&0.3210 $\pm$ 0.005&0.4086 $\pm$ 0.006 \\
S& 0.5506 $\pm$ 0.004&0.5685 $\pm$ 0.004&0.5392 $\pm$ 0.004&0.5952 $\pm$ 0.004 \\
T& 0.1514 $\pm$ 0.001&0.2258 $\pm$ 0.008&0.1791 $\pm$ 0.001&0.3030 $\pm$ 0.001 \\
\midrule
P+ A& 0.8730 $\pm$ 0.001&0.8835 $\pm$ 0.002&0.8647 $\pm$ 0.002&0.8915 $\pm$ 0.002 \\
P+ S& 0.8877 $\pm$ 0.003&0.9065 $\pm$ 0.003&\textbf{0.8722 $\pm$ 0.003}&0.9017 $\pm$ 0.002 \\
P+ T& 0.8651 $\pm$ 0.003&0.8712 $\pm$ 0.003&0.8613 $\pm$ 0.004&0.8847 $\pm$ 0.002 \\
A+ S& 0.6368 $\pm$ 0.006&0.6642 $\pm$ 0.008&0.6203 $\pm$ 0.006&0.6789 $\pm$ 0.005 \\
A+ T& 0.3286 $\pm$ 0.008&0.3584 $\pm$ 0.010&0.3176 $\pm$ 0.007&0.4042 $\pm$ 0.005 \\
S+ T& 0.5567 $\pm$ 0.004&0.5765 $\pm$ 0.005&0.5446 $\pm$ 0.004&0.6062 $\pm$ 0.004 \\
\midrule
P+ A+ S&\textbf{ 0.8883 $\pm$ 0.003}&\textbf{0.9105 $\pm$ 0.003}&0.8704 $\pm$ 0.006&\textbf{0.9022 $\pm$ 0.003} \\
P+ A+ T& 0.8713 $\pm$ 0.003&0.8806 $\pm$ 0.004&0.8643 $\pm$ 0.003&0.8901 $\pm$ 0.003 \\
P+ S+ T& 0.8822 $\pm$ 0.005&0.9024 $\pm$ 0.005&0.8656 $\pm$ 0.006&0.8974 $\pm$ 0.004 \\
A+ S+ T& 0.6322 $\pm$ 0.004&0.6657 $\pm$ 0.006&0.6130 $\pm$ 0.003&0.6784 $\pm$ 0.004 \\
\midrule
P+ A+ S+ T& 0.8829 $\pm$ 0.003&0.9034 $\pm$ 0.003&0.8662 $\pm$ 0.005&0.8978 $\pm$ 0.003 \\

\bottomrule
\end{tabular}}
  \vspace{-9pt}
\end{table}

\begin{table}[h!]
\caption{Ablation Study on the Impact of Different Latent Features Combined with Preprocessed Features on Model Performance}
\vspace{-3px}
\centering
\label{tab:abalation_feature_latent}
\resizebox{\columnwidth}{!}{\begin{tabular}{c|c|c|c|c|c}
\toprule

\textbf{Features}&\textbf{Latent F.} &\textbf{F1 Score} &\textbf{Precision}&\textbf{Recall} &\textbf{Accucary} \\
\midrule

P+ A+ S & P& 0.9148 $\pm$ 0.006&0.9212 $\pm$ 0.008&0.8910 $\pm$ 0.009&0.9045 $\pm$ 0.008 \\
P+ A+ S & A& \underline{0.9189 $\pm$ 0.002}&\underline{0.9274 $\pm$ 0.003}&\textbf{0.9004 $\pm$ 0.002}&\textbf{0.9124 $\pm$ 0.003} \\
P+ A+ S & S& 0.9190 $\pm$ 0.003&0.9257 $\pm$ 0.005&0.8971 $\pm$ 0.005&0.9108 $\pm$ 0.004 \\
P+ A+ S & T& 0.9146 $\pm$ 0.001&0.9233 $\pm$ 0.003&0.8923 $\pm$ 0.002&0.9062 $\pm$ 0.002 \\
\midrule
P+ A+ S & P+ A& 0.9163 $\pm$ 0.002&0.9253 $\pm$ 0.002&0.8932 $\pm$ 0.003&0.9074 $\pm$ 0.002 \\
P+ A+ S & P+ S& 0.9175 $\pm$ 0.002&0.9266 $\pm$ 0.002&0.8960 $\pm$ 0.004&0.9096 $\pm$ 0.003 \\
P+ A+ S & P+ T& 0.9152 $\pm$ 0.001&0.9233 $\pm$ 0.002&0.8922 $\pm$ 0.002&0.9061 $\pm$ 0.001 \\
P+ A+ S & A+ S& 0.9153 $\pm$ 0.001&0.9241 $\pm$ 0.003&0.8937 $\pm$ 0.002&0.9073 $\pm$ 0.003 \\
P+ A+ S & A+ T& 0.9124 $\pm$ 0.003&0.9197 $\pm$ 0.004&0.8911 $\pm$ 0.005&0.9038 $\pm$ 0.004 \\
P+ A+ S & S+ T& 0.9133 $\pm$ 0.003&0.9200 $\pm$ 
0.004&0.8901 $\pm$ 0.004&0.9034 $\pm$ 0.004 \\
\midrule
P+ A+ S & P+ A+ S& \textbf{0.9193 $\pm$ 0.003}&\textbf{0.9301 $\pm$ 0.003}&\underline{0.8990 $\pm$ 0.004}&\underline{0.9118 $\pm$ 0.003} \\
P+ A+ S & P+ A+ T& 0.9151 $\pm$ 0.001&0.9241 $\pm$ 0.002&0.8914 $\pm$ 0.001&0.9060 $\pm$ 0.001 \\
P+ A+ S & P+ S+ T& 0.9149 $\pm$ 0.001&0.9219 $\pm$ 0.002&0.8918 $\pm$ 0.003&0.9053 $\pm$ 0.003 \\
P+ A+ S & A+ S+ T& 0.9123 $\pm$ 0.004&0.9198 $\pm$ 0.004&0.8898 $\pm$ 0.005&0.9032 $\pm$ 0.005 \\
\midrule
P+ A+ S&-& 0.8883 $\pm$ 0.003&0.9105 $\pm$ 0.003&0.8704 $\pm$ 0.006&0.9022 $\pm$ 0.003 \\
-& P+ A+ S+ T& 0.3041 $\pm$ 0.001&0.1949 $\pm$ 0.013&0.1779 $\pm$ 0.000&0.1454 $\pm$ 0.002 \\
P+ A+ S & P+ A+ S+ T& 0.9126 $\pm$ 0.004&0.9203 $\pm$ 0.004&0.8868 $\pm$ 0.004&0.9017 $\pm$ 0.004 \\

\midrule
\bottomrule
\end{tabular}}
\vspace{-4px}
\end{table}

In addition, we assessed the impact of various feature combinations on our model's performance (Table. \ref{tab:abalation_feature_latent}). We found that latent application usage features alone (A) yielded the highest accuracy (0.9274) and recall (0.9004), underscoring their predictive power for occupation inference. However, the combination of physical, application, and social \& environmental features (P+A+S) provided the most balanced performance across all metrics, achieving the best F1-score (0.9193) and precision (0.9301). Including temporal information generally resulted in lower performance, suggesting its limited relevance. Further analysis highlighted that using only preprocessed or latent features reduces effectiveness. Models with only preprocessed features (P+A+S) saw diminished performance, likely due to missing nuanced patterns captured by latent features. Conversely, models relying solely on latent features (P+A+S+T) performed worse, as these features, while capturing underlying data structures, lack the immediate informativeness of preprocessed data. This demonstrates the need to integrate both processed and latent features for optimal performance.

\vspace{-2px}
\subsection{Implications and Limitations }

By providing tailored recommendations, digital assistants that accurately infer occupations can significantly enhance task management. For example, if the system recognizes a user as a student, it could prioritize academic scheduling and resource suggestions. Conversely, the assistant might focus on customer management and sales tracking tasks if the user is identified as a retail worker.

Despite its potential, this study has limitations that affect its broader applicability and raise privacy concerns. The current dataset is relatively small and not diverse enough across different occupational categories, which may limit the generalizability of our findings. Furthermore, inconsistencies in sensor data collection led to the exclusion of about 2,000 instances, often because some sensors failed to collect any data. To overcome these issues, our future work will explore machine learning models capable of handling missing data. This approach will allow us to utilize incomplete datasets more effectively. It has not yet been implemented but is planned for further development. Additionally, we aim to expand the dataset and standardize data collection across different mobile devices to improve the model’s accuracy and utility in real-world applications. Privacy remains a major concern as the use of personal data for training models requires strict adherence to data protection standards. Future initiatives will explore federated learning to learn from decentralized data without compromising personal information. We also plan to explore the potential of synthetic data generation to mitigate privacy risks further, ensuring our systems are both effective and comply with ethical standards.

\vspace{-2px}
\section{Conclusion}

We presented the problem of occupation inference through pervasive signals associated with different tasks, crucial for enhancing digital assistants. Our WorkR model uses passively captured sensor data and latent features to infer occupations accurately, demonstrating significant potential for personalized and proactive support. We discussed the implications of occupation inference for personalizing task support and recommendations in digital assistants. Importantly, future work will explore enhancing data privacy measures, ensuring that our approach aligns with ethical standards and privacy regulations. This research opens avenues for advanced task support, management, and tailored recommendations for task progression and completion, highlighting the transformative potential of occupation-aware technologies.
\vspace{-2px}
\section*{Acknowledgement}
This research received partial support from Microsoft Research, which provided the dataset through the Microsoft-RMIT Cortana Intelligence Institute, and the ARC Centre of Excellence for Automated Decision-Making and Society (ADM+S).

\newpage



\bibliographystyle{ACM-Reference-Format}
\bibliography{mybibfile}


\end{document}